\documentclass[lettersize,journal]{IEEEtran}
\usepackage{amsmath,amsfonts}
\usepackage{array}
\usepackage[caption=false,font=normalsize,labelfont=sf,textfont=sf]{subfig}
\usepackage{textcomp}
\usepackage{stfloats}
\usepackage{url}
\usepackage{verbatim}
\usepackage{graphicx}
\def\BibTeX{{\rm B\kern-.05em{\sc i\kern-.025em b}\kern-.08em
    T\kern-.1667em\lower.7ex\hbox{E}\kern-.125emX}}
\usepackage{balance}

\usepackage{algorithm}
\usepackage{booktabs}
\usepackage{multirow}
\usepackage{algorithm}
\usepackage{algpseudocode}

\usepackage{hyperref}
\hypersetup{
    hidelinks
}
\usepackage{xcolor}
\definecolor{orcidlogocol}{HTML}{A6CE39}
\usepackage{orcidlink}

\begin{document}

\markboth{%
	\parbox[c]{0.88\textwidth}{%
		\raggedright
		This work has been submitted to the IEEE for possible publication.\\
		Copyright may be transferred without notice, after which this version may no longer be accessible.
	}%
}{}

\title{Industrial Agentic AI: LLM-Driven New-Generation Agents for Smart Manufacturing}

\author{
	\IEEEauthorblockN{
		Lei Ren\textsuperscript{1,*},
		Haiteng Wang\textsuperscript{2},
		Shixiang Li\textsuperscript{1},
		Jing Zhang\textsuperscript{1},
		Yuqing Wang\textsuperscript{1},
		Jingwei Guo\textsuperscript{1},
		Yunfei Zhu\textsuperscript{3},
		Yuanjun Laili\textsuperscript{1},
		Yang Tang\textsuperscript{4},
		Lihui Wang\textsuperscript{5}
	}\\
	\IEEEauthorblockA{
		\textsuperscript{1}School of Automation Science and
		Electrical Engineering, Beihang University,
		Beijing 100191, China\\
		\textsuperscript{2}Institute of Artificial Intelligence,
		Beihang University, Beijing 100191, China\\
		\textsuperscript{3}School of Software, Beihang University,
		Beijing 100191, China\\
		\textsuperscript{4}Key Laboratory of Smart Manufacturing
		in Energy Chemical Process, Ministry of Education,
		East China University of Science and Technology,
		Shanghai 200237, China\\
		\textsuperscript{5}Division of Production Systems and
		Automation, School of ITM, KTH Royal Institute of
		Technology, 10044 Stockholm, Sweden\\
		\textsuperscript{*}Corresponding author:
		renlei@buaa.edu.cn
	}
}

\maketitle
\begin{abstract}
Large language models (LLMs) are enabling agents to move beyond predefined workflows toward autonomous planning and execution, creating new opportunities for smart manufacturing. However, dynamic environments, complex physical constraints, and stringent reliability requirements hinder the direct deployment of general-purpose agents in smart manufacturing. This article presents a systematic overview of LLM-driven industrial agents. First, we analyze five key challenges and discuss the concept and core characteristics of industrial agents. We then establish a unified taxonomy covering application scenarios, embodiment forms, and five capability levels. Next, we propose a four-layer system architecture and detail its core modules and operational mechanisms for closed-loop task execution. Finally, we review representative applications across the industrial lifecycle and outline future research directions. This article provides a structured reference for the design, development, and application of industrial agents.

\end{abstract} 

\begin{IEEEkeywords}
LLM, Industrial Agents, Agentic AI, Multi-Agent System, Industrial foundation model.
\end{IEEEkeywords}

\section{Introduction}
\IEEEPARstart{A}{gent}  \cite{russell1995artificial,wooldridge1995intelligent, shi2020survey} is commonly defined as a computational entity capable of perceiving its environment, making autonomous decisions, and taking actions accordingly. Agent theory has evolved over several decades, with extensive industrial applications such as multi-agent-enabled unmanned aerial vehicle coordination\cite{jiang2021agent,jia2026aoi} and distributed power-grid scheduling \cite{foerster2016learning,lowe2017multiagent,rashid2018qmix}. Since 2023, large language models (LLMs) has ushered agents into a new stage of development \cite{yao2023react,schick2023toolformer,shinn2023reflexion}. Recent agents, such as Claude Code, Codex, and OpenClaw, have demonstrated strong capabilities in context understanding, autonomous decision-making, and task execution. Meanwhile, protocols such as Agent-to-Agent (A2A) and the Model Context Protocol (MCP) have improved cross-platform interoperability among agents \cite{google2025a2a,anthropic2024mcp}. These advances are driving the shift from conventional agents to LLM-driven agents with continuous perception, planning, execution, and collaboration \cite{sapkota2026aiagents}.

LLM-driven industrial agents provide new opportunities for industrial artificial intelligence \cite{ren2024deeplearning}. However, current industrial applications remain at an early stage and often rely on predefined workflows. Although several studies \cite{wang2026collm, shen2006applications,leitao2009agentbased} have applied multi-agent systems to tasks such as manufacturing scheduling and equipment fault diagnosis, existing methods are still limited in core industrial tasks such as process decision-making and control. These tasks involve dynamic environments, strict accuracy, and therefore require agents to understand industrial mechanisms and process logic while ensuring reliable execution \cite{ren2025industrial}. In particular, industrial agents face the following major challenges.

\textit{1) Lack of industrial domain knowledge and multimodal data understanding.} General-purpose agents are primarily trained on general-domain corpora, lacking  specialized knowledge and mechanisms of industrial processes. They also struggle to interpret industrial multimodal data and their relationships. These limitations hinder their ability to accurately interpret industrial operating states and understand their physical significance.

\textit{2) Insufficient reliability in autonomous industrial decision-making.} General-purpose agents may tolerate uncertainty or errors in decision-making. Industrial applications require much higher reliability, especially in robotic manipulation and equipment fault diagnosis. Existing agents are largely built on probabilistic AI models, whose stochastic outputs can lead to unreliable decisions\cite{nikitin2024kernel}. Without reliable verification before execution, such errors can lead to equipment damage or production disruptions \cite{cen2024feasibility,wallace2024physics}.

\textit{3) Limited self-organized collaboration among industrial agents.} Agents are effective for relatively independent tasks with predefined interaction rules.
However, industrial systems require close coordination across coupled processes, such as production scheduling, quality control, and supply chains. Agents may differ in objectives, data, capabilities, leading to inconsistent information, communication bottlenecks, and decision conflicts \cite{ding2024magi,sun2024t2mac}. Existing agents still lack sufficient global process awareness and self-organizing mechanisms for dynamic coordination\cite{bo2024reflective}.

\textit{4) Limited ability to execute long-horizon industrial tasks.} Industrial tasks involve tightly coupled stages, where later operations depend on the physical outcomes of earlier ones. Existing agents struggle to track these dependencies and revise plans throughout execution, allowing local errors to propagate and disrupt the overall workflow. \cite{nayak2024longhorizon,yao2023react,schick2023toolformer,shinn2023reflexion}. 

\textit{5) Weak adaptability to changing industrial environments} General-purpose agents are often designed for static settings with clear task boundaries. Industrial environments are highly dynamic, with frequent changes in equipment conditions, and resource constraints, requiring agents to continually adjust their behavior \cite{ren2025foundation}. However, existing agents still lack continual adaptation capabilities under industrial data drift and scenario changes. \cite{galashov2024nonstationary,lee2024pausing}.

These challenges indicate that existing general-purpose agents remain insufficient for high-reliability industrial applications. Industrial agents are therefore not a simple combination of general-purpose agents and industrial software or hardware. Instead, Agentic AI calls for a systematic reconsideration of their theoretical foundations, system architectures, and implementation methodologies. Despite the rapidly growing body of related research \cite{yao2023react,schick2023toolformer,shinn2023reflexion}, a systematic study of next-generation industrial agents is still lacking. This article therefore investigates the definition, characteristics, system architecture, enabling technologies, and representative applications of industrial agents, and further discusses their future research directions. The main contributions are summarized as follows:

1) Five Key challenges are analyzed in applying agents to industrial scenarios, including industrial knowledge understanding, decision reliability, self-organized collaboration, long-horizon task execution, and  industrial environments self-adaptation.

2) A conceptual description of industrial agents is provided, and a unified taxonomy encompassing application scenarios, embodiment forms, and a five-level capability hierarchy is established.

3) A system architecture for industrial agent is proposed, detailing its core modules and their operational mechanisms for closed-loop task execution.

4) Representative applications across the industrial lifecycle are provided, and future research opportunities and open issues are discussed.

\section{Evolution of Industrial Agents}

\subsection{Conception of Industrial Agents}
Industrial agents are a new type of intelligent agent system driven by emerging artificial intelligence technologies, such as foundation models and embodied intelligence, serving applications throughout the lifecycle of industrial products. They include both embodied agents\cite{ren2024embodied} and non-embodied forms (software-based systems), and possess core capabilities including industrial perception and understanding, knowledge memory, planning and decision-making, communication and interaction, autonomous execution, and evolutionary learning. These capabilities enable intelligent services throughout the full industrial business chain, covering R\&D design, manufacturing, testing and validation, business management, and operation and maintenance.
\subsection{Five Characteristics of New-Generation Industrial Agents}
\begin{figure*}[htbp]
    \centering
    \includegraphics[width=0.8\textwidth]{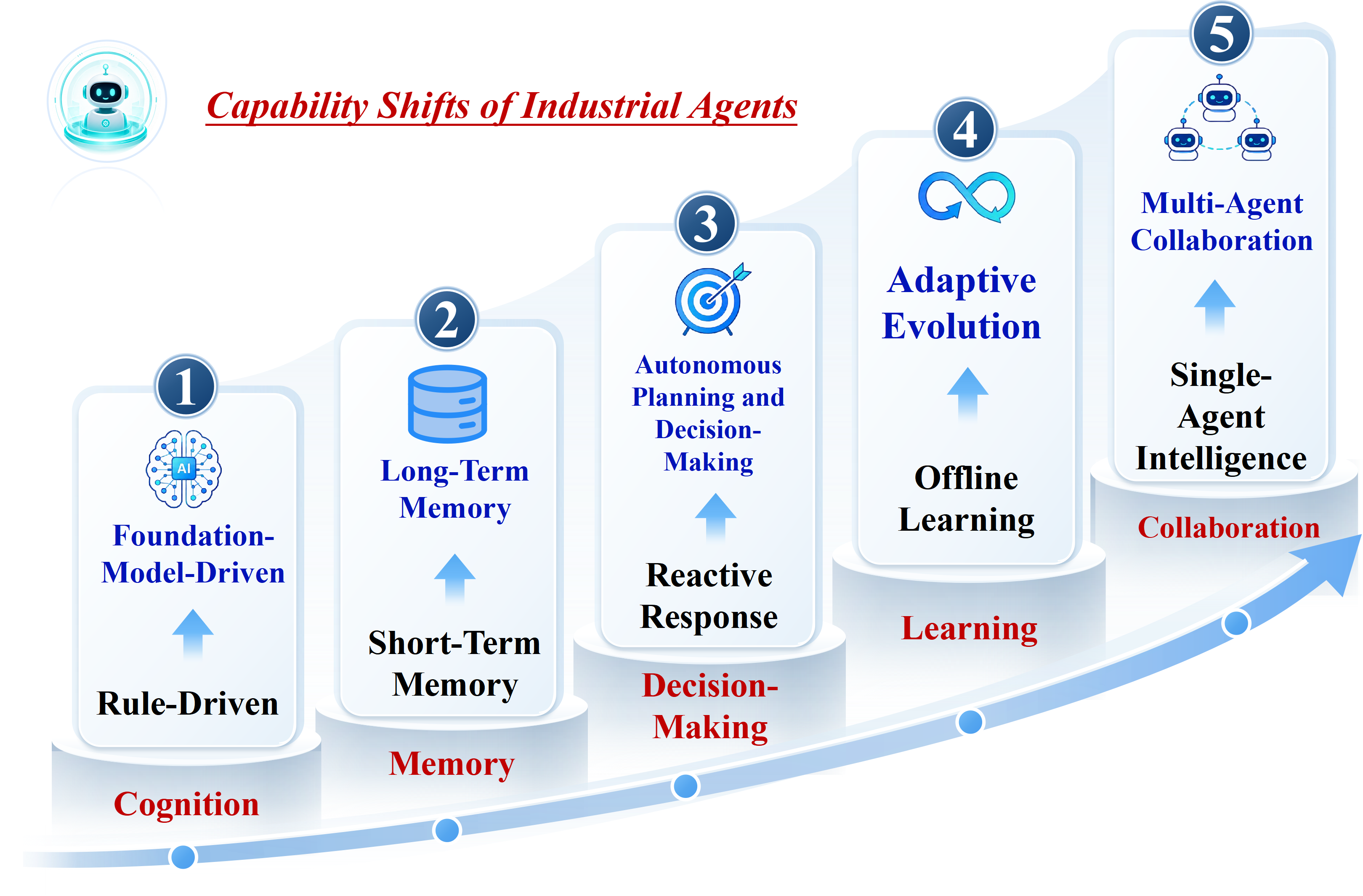}
    \caption{Core Capability Shifts of Industrial Agents}
    \label{fig1}
\end{figure*}

The core characteristics of industrial agents reflect their capability evolution toward handling complex industrial tasks. Compared with traditional industrial intelligent systems that rely on fixed rules, limited information, and predefined workflows, industrial agents exhibit systematic capability shifts in their cognition, memory, decision-making, learning, and collaboration. These advancements are specifically manifested as the transition from rule-driven intelligence to foundation-model-driven intelligence, from short-term memory to long-term memory, from reactive response to autonomous planning and decision-making, from offline learning to adaptive evolution, and from single-agent intelligence to multi-agent collaboration. These evolutionary processes collectively enable industrial agents to perform complex task understanding, continuous decision-making, closed-loop execution, adaptive learning, and collaborative task execution in dynamic industrial environments, as illustrated in Fig.~\ref{fig1}.

\subsubsection{From Rule-Driven Intelligence to Foundation-Model-Driven Intelligence}
Foundation-model-driven intelligence represents the transition of industrial agents from predefined rule execution to model-driven task understanding, reasoning, and solution generation. By leveraging industrial data training and domain knowledge enhancement, industrial agents can understand task objectives and operational conditions, and generate executable solutions by incorporating industrial mechanisms and professional tools. Compared with traditional rule-based systems, this paradigm provides stronger adaptability to complex, dynamic, and uncertain industrial environments, supporting tasks such as production scheduling, fault handling, and process planning, while ensuring decision reliability and feasibility through industrial constraints.

\subsubsection{From Short-Term Memory to Long-Term Memory}
Long-term memory represents the transition of industrial agents from local context processing to continuous knowledge accumulation. By persistently storing and associating key states, historical experiences, and decision outcomes throughout task execution, industrial agents can maintain cognitive continuity in long-cycle industrial tasks and utilize accumulated experience to support current decisions. Compared with short-term memory that relies only on temporary contexts, long-term memory is more suitable for tasks requiring historical information, such as fault tracing, equipment health management, and process change analysis, thereby improving decision consistency, traceability, and experience reuse capability.

\subsubsection{From Reactive Response to Autonomous Planning and Decision-Making}
Autonomous planning and decision-making represent the transition of industrial agents from event-triggered local responses to goal-oriented proactive decision-making. Industrial agents can analyze current tasks, operational states, and constraints to autonomously perform task decomposition, solution generation, and execution adjustment, while dynamically re-planning based on process feedback. Compared with traditional reactive response approaches, autonomous planning and decision-making are more suitable for industrial scenarios with frequent disturbances, complex task dependencies, and strong resource constraints, enabling applications such as production scheduling, test organization, maintenance planning, and supply chain coordination, and improving continuous decision-making and closed-loop execution capabilities.

\subsubsection{From Offline Learning to Adaptive Evolution}
Adaptive evolution represents the transition of industrial agents from static capability deployment to continuous learning and optimization. Industrial agents can identify capability deviations according to environmental changes, task states, and execution feedback, and continuously refine their knowledge, memory, decision strategies, and execution behaviors through operational feedback and capability updating. Compared with offline learning methods that remain fixed after initial training, adaptive evolution is more suitable for dynamic industrial scenarios involving equipment degradation, process adjustments, and production condition variations. It supports tasks such as predictive maintenance, process optimization, and quality control, enabling industrial agents to maintain stable adaptability and task performance during long-term operation.

\subsubsection{From Single-Agent Intelligence to Multi-Agent Collaboration}
Multi-agent collaboration represents a shift in the collaboration paradigm of industrial agents, from isolated single-agent task solving to coordinated execution and system-level collaboration. Multiple agents can perform role allocation, task orchestration, and information interaction around complex industrial objectives, and collaboratively accomplish cross-process tasks through negotiation and feedback. Compared with single-agent intelligence, multi-agent collaboration is more suitable for complex scenarios such as cross-process production scheduling, process-quality collaborative optimization, equipment maintenance coordination, and supply chain management, improving task parallelism, professional specialization, and system-level decision-making capability.

\section{Taxonomy of Industrial Agents}
\begin{figure*}[htbp]
    \centering
    \includegraphics[width=0.9\textwidth]{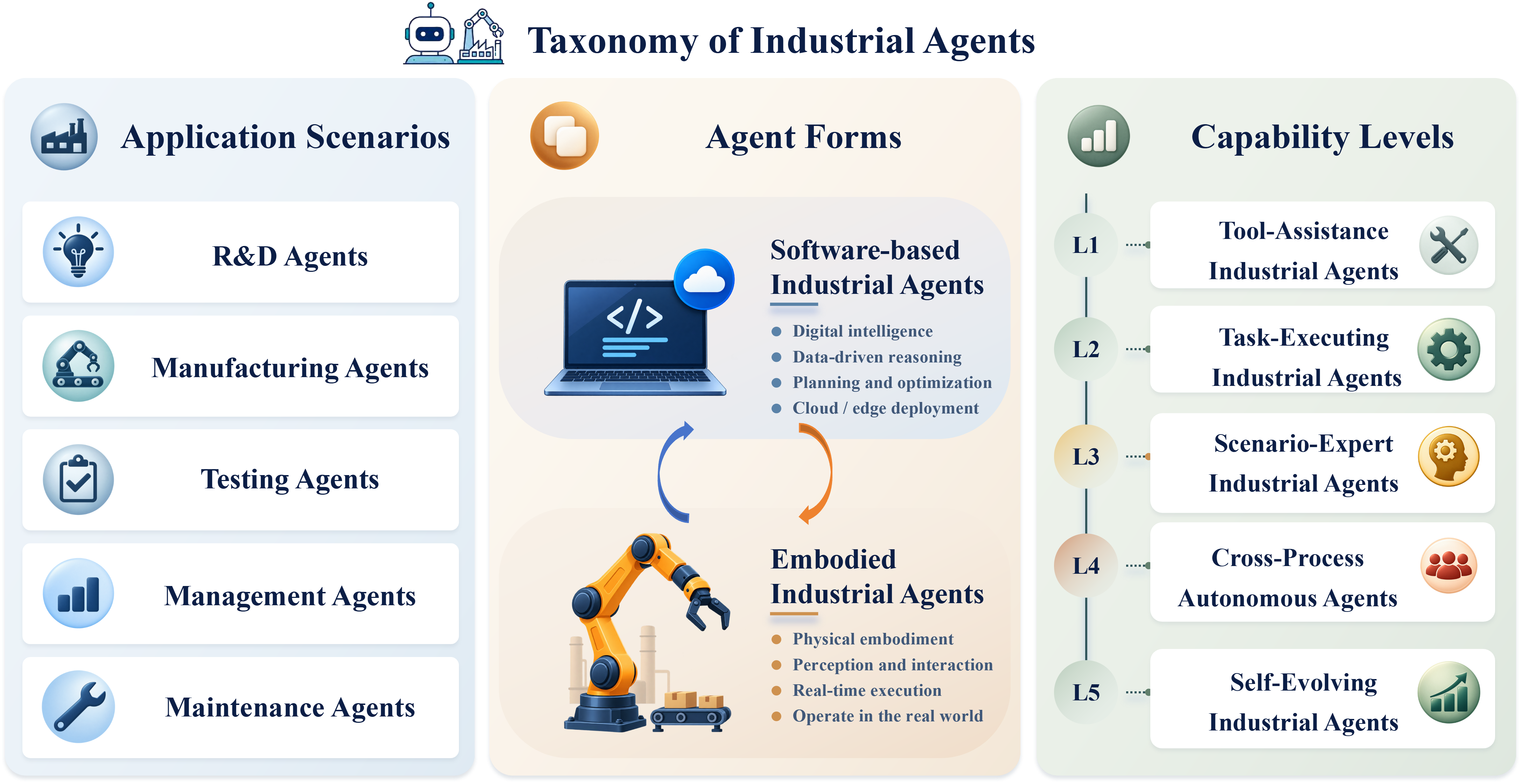}
    \caption{Taxonomy of Industrial Agents}
    \label{fig3}
\end{figure*}

Industrial agents vary considerably in their application objectives, implementation forms, and degrees of autonomy. To distinguish these differences, industrial agents are classified from three complementary dimensions: application scenarios, agent forms, and capability levels. The application-scenario dimension describes where industrial agents are deployed and what types of industrial activities they support; the agent-form dimension distinguishes whether their intelligence is primarily realized in digital systems or directly embodied in the physical world; and the capability-level dimension reflects their progression from tool-assisted operation to collaborative autonomy and continuous self-evolution. The resulting taxonomy is illustrated in Fig.~\ref{fig3}.

\subsection{Classification by Application Scenarios}

From the perspective of application scenarios, industrial agents can be classified according to the major business activities across the industrial product lifecycle. Although the data sources, professional knowledge, software tools, and execution environments differ among scenarios, the underlying objective is consistent: to combine industrial intelligence with domain tasks and improve the efficiency, quality, and autonomy of industrial operations.

\subsubsection{R\&D Agents}

R\&D agents support activities such as product design, engineering analysis, simulation, and process planning. They interpret design requirements and engineering constraints, organize relevant domain knowledge, and interact with professional software to assist engineers in generating and evaluating design solutions. Typical tools include CAD, CAE, CAM, optimization solvers, and simulation platforms. In practice, these agents can participate in structural design, performance evaluation, parameter optimization, and process planning, reducing repetitive operations across different engineering tools and improving the continuity of the design process.

\subsubsection{Manufacturing Agents}

Manufacturing agents operate around production planning and shop-floor execution. They integrate information from production orders, equipment states, materials, process conditions, and quality data to support scheduling, process adjustment, equipment coordination, and disturbance handling. Through connections with MES, MOM, industrial control systems, and production equipment, these agents can assist or autonomously perform production scheduling, process parameter optimization, resource allocation, and equipment coordination. Their role is particularly important in manufacturing environments where production states and resource conditions change frequently.

\subsubsection{Testing Agents}

Testing agents are designed for test planning, test execution, monitoring, data analysis, and result evaluation. They interpret testing standards, experimental requirements, and evaluation criteria, and organize appropriate testing procedures according to the object under test and its operating state. By interacting with testing equipment, simulation platforms, data-analysis tools, and test-management systems, these agents can support parameter configuration, process monitoring, anomaly identification, and result assessment. This reduces repetitive manual operations and improves the consistency of industrial testing processes.

\subsubsection{Management Agents}

Management agents focus on operational and business-level decision support, including production planning, resource allocation, demand forecasting, inventory management, production-sales coordination, and supply-chain management. They integrate data from enterprise information systems such as ERP, SCM, and CRM to identify operating conditions, analyze potential risks, and generate decision alternatives. Compared with conventional information systems that mainly provide data recording and reporting functions, management industrial agents place greater emphasis on continuous analysis, decision support, and cross-system coordination.

\subsubsection{Maintenance Agents}

Maintenance agents support equipment monitoring, fault diagnosis, root-cause analysis, maintenance planning, and service management. They combine operational data, alarm information, maintenance records, technical documents, and historical fault cases to identify abnormal states and assist in determining appropriate maintenance actions. By interacting with PHM platforms, maintenance-management systems, spare-parts systems, and diagnostic tools, these agents can support fault localization, repair-plan generation, maintenance scheduling, and remote technical services. Their use extends maintenance activities from passive fault handling toward condition-based and predictive maintenance.

\subsection{Classification by Agent Forms}

According to their implementation forms and modes of interaction with industrial environments, industrial agents can be divided into two basic forms: software-based industrial agents and embodied industrial agents. These two forms are not isolated from each other. Software-based agents provide flexible digital intelligence for knowledge processing, reasoning, planning, and optimization, whereas embodied agents extend these capabilities into the physical world through sensing, control, and real-time interaction. In practical industrial systems, the two forms can cooperate and complement each other to bridge digital and physical environments.

\subsubsection{Software-based Industrial Agents}

Software-based industrial agents primarily operate in digital environments and use software systems as their main carriers. Their intelligence is realized through industrial data, domain knowledge, foundation models, professional algorithms, and industrial software. These agents can access databases, knowledge bases, APIs, optimization models, and engineering software to perform information understanding, data-driven reasoning, planning, and decision support.

Typical software-based industrial agents can be integrated with ERP, MES, MOM, PHM, CAD, CAE, and other industrial systems. Depending on the application, they may be deployed locally, at the edge, or in cloud environments. Their main advantages lie in flexible deployment, convenient integration with existing information systems, and relatively low cost for capability extension. They are therefore suitable for a broad range of tasks such as design assistance, production scheduling, quality analysis, fault diagnosis, and operational decision-making.

\subsubsection{Embodied Industrial Agents}

Embodied industrial agents possess physical carriers and directly interact with the industrial environment. Typical carriers include industrial robots, mobile robots, AGVs, intelligent machine tools, inspection equipment, and other cyber-physical devices. Through cameras, force sensors, encoders, industrial sensors, and other perception devices, these agents acquire information from the physical environment and translate planning and decision results into real-time actions.

Compared with software-based agents, embodied industrial agents place greater emphasis on physical embodiment, environmental perception, real-time execution, and safe interaction. They can perform tasks such as robotic assembly, material handling, equipment inspection, automated testing, and on-site maintenance. Since their actions directly affect physical equipment and production processes, their deployment requires stricter guarantees in terms of real-time performance, control accuracy, operational safety, and robustness to environmental disturbances.

\subsection{Classification by Capability Level}

The capability level reflects the degree of autonomy and complexity that an industrial agent can handle. Rather than distinguishing agents only by the models or tools they use, the classification focuses on what they can accomplish in industrial tasks. The five levels progress from tool assistance to task execution, scenario expertise, cross-process autonomy, and self-evolution.

\subsubsection{Level 1: Tool-Assistance Industrial Agents}

Tool-assistance industrial agents mainly serve as intelligent assistants for human operators. Their functions typically include information retrieval, data organization, document generation, basic analysis, parameter calculation, and invocation of professional tools according to explicit instructions. Human users remain responsible for defining task objectives, selecting key procedures, and making final decisions.

Typical applications include technical-document retrieval, production-data summarization, report generation, standard calculations, and software-assisted engineering analysis. At this stage, the agent improves work efficiency but does not independently control the overall task process.

\subsubsection{Level 2: Task-Executing Industrial Agents}

Task-executing industrial agents can independently complete well-defined industrial tasks within explicit objectives, procedures, and operational boundaries. They are capable of interpreting task requirements, organizing multiple execution steps, invoking appropriate tools, and producing structured outputs without continuous human intervention.

Such agents can be used for standardized data processing, routine fault diagnosis, process-parameter analysis, test-report generation, and workflow execution. Their autonomy, however, remains bounded by predefined task scopes. When unexpected conditions or situations outside these boundaries occur, human intervention or higher-level decision support is still required.

\subsubsection{Level 3: Scenario-Expert Industrial Agents}

Scenario-expert industrial agents are oriented toward relatively complete industrial scenarios rather than isolated tasks. They combine domain knowledge, industrial mechanisms, historical experience, professional tools, and contextual information to analyze problems and generate solutions within a specific application domain.

Compared with task-executing agents, they can deal with stronger task dependencies and more complex constraints, and are capable of task decomposition, solution generation, execution planning, and adjustment according to feedback. Representative applications include complex-equipment fault diagnosis, maintenance decision-making, process planning, test analysis, and production scheduling. Human confirmation is generally retained for safety-critical operations and decisions with significant operational consequences.

\subsubsection{Level 4: Cross-Process Autonomous Agents}

Cross-process autonomous agents extend autonomy from a single task or scenario to coordinated operation across multiple processes, equipment, and information systems. They can coordinate task dependencies, information exchange, resource allocation, and execution across interconnected industrial processes according to shared operational objectives.

This level is suitable for tasks that cannot be effectively handled by an isolated agent, such as cross-line production scheduling, coordinated equipment maintenance, production-quality joint optimization, complex test organization, and supply-chain coordination. The emphasis is not merely on autonomous execution of individual tasks, but on maintaining coordinated and closed-loop operation across interconnected industrial processes and systems. Human supervision remains necessary for major failures, high-risk operations, and situations beyond the authorized operating envelope.

\subsubsection{Level 5: Self-Evolving Industrial Agents}

Self-evolving industrial agents represent a higher level of capability in which agents can continuously improve their own knowledge, strategies, skills, and collaboration mechanisms according to long-term operational experience and environmental changes. Instead of relying solely on capabilities fixed at deployment time, these agents can identify performance gaps, accumulate reusable experience, adjust decision strategies, and refine task-execution procedures through feedback.

Self-evolving industrial agents are intended for open and dynamic industrial environments in which equipment conditions, process requirements, task objectives, and available resources may change over time. Their evolution should nevertheless remain subject to explicit industrial constraints, safety requirements, validation procedures, and human governance. The objective is therefore not unrestricted autonomous modification, but controlled capability evolution that allows industrial agents to remain effective as operating conditions and industrial knowledge continue to change.

\section{Architecture}
\begin{figure*}[htbp]
    \centering
    \includegraphics[width=\textwidth]{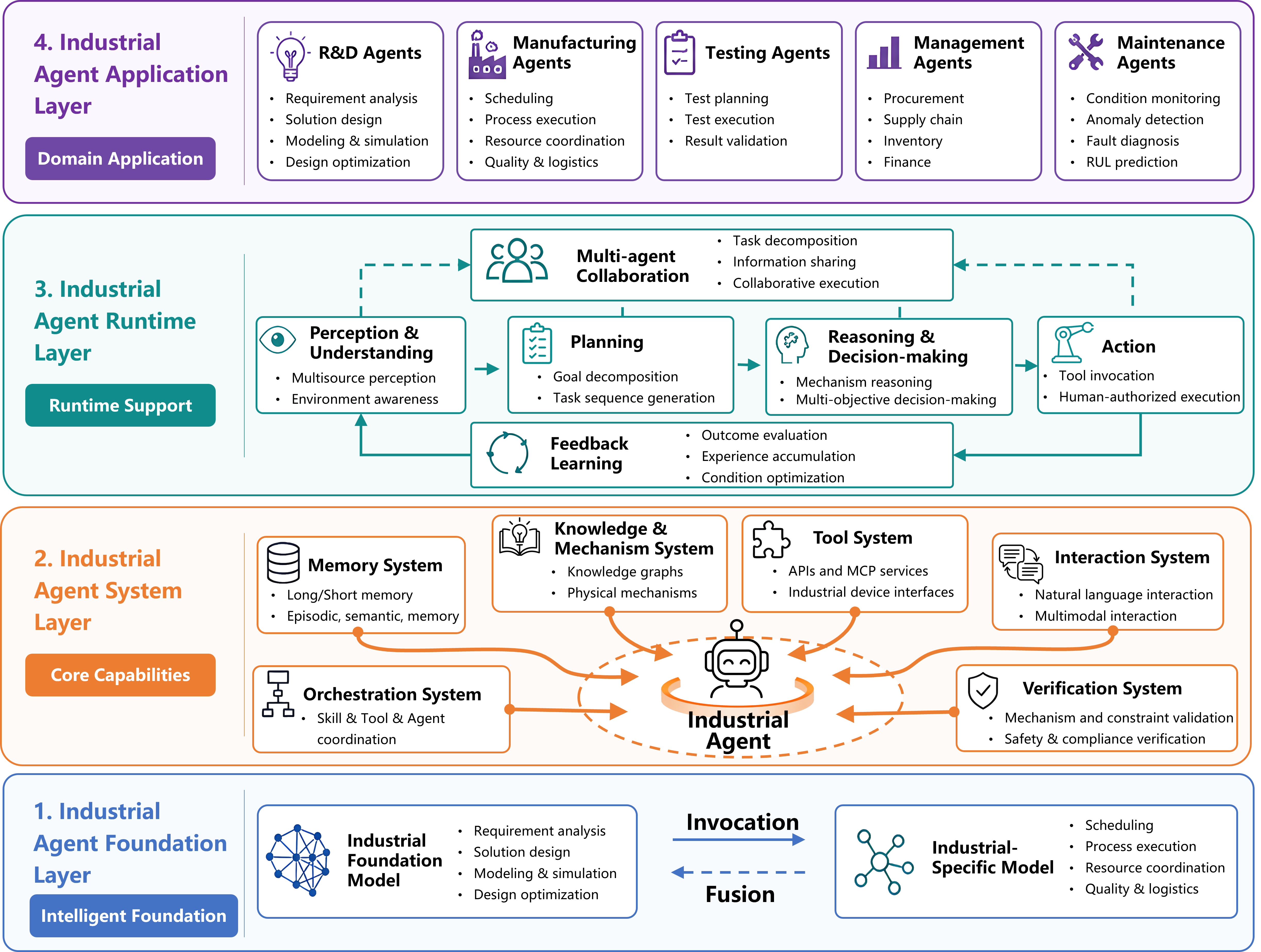}
    \caption{Internal Enhancement of Industrial Agents}
    \label{fig4_1}
\end{figure*}

This paper divides the industrial agent architecture into four layers, as shown in Fig. ~\ref{fig4_1}. The Industrial Agent Foundation Layer provides the industrial foundation model and industrial-specific models. The Industrial Agent System Layer defines the agent's core components: memory, knowledge and mechanisms, tools, interaction, orchestration, and verification. The Industrial Agent Runtime Layer describes how these components support perception and understanding, planning, reasoning and decision-making, action, and feedback learning, with multi-agent collaboration across the process. The Industrial Agent Application Layer covers R\&D, manufacturing, testing, management, and maintenance. The system layer specifies the agent's composition, while the runtime layer describes its operation.

\subsection{Industrial Agent Foundation Layer}
The Industrial Agent Foundation Layer is built around industrial foundation models, complemented by industrial specialized models and industrial knowledge and mechanisms that provide specialized capabilities and domain-specific priors. Industrial foundation models are responsible for multi-source information understanding, task reasoning, planning \& decision-making, and tools invocation. Industrial specialized models provide capabilities for specialized analysis, process prediction, simulation, and optimization. These components together provide the fundamental intelligence required for industrial agents to understand and interact with industrial objects and processes.

\subsubsection{Industrial Foundation Models}
The industrial foundation model supports task understanding and reasoning in industrial agents. It is trained on industrial data and adapted to specific domains. Multimodal alignment links information from text, images, time-series, and engineering drawings to equipment, components, processes, operating conditions, and production tasks. For a given task, the model draws on specialized models, industrial knowledge, and physical mechanisms to interpret the current state, develop plans, and make decisions.

\subsubsection{Industrial Specialized Models}
Industrial specialized models include industrial world models, task-specific models, and mechanism-based models. Industrial world models predict how system states change under different actions and operating conditions. Depending on their training scope, they can describe process dynamics, performance degradation, and fault propagation. Task-specific models address tasks such as time-series analysis, fault diagnosis, and process optimization. Mechanism-based models use physical equations to describe equipment behavior and process relationships. The industrial foundation model invokes these models through standardized interfaces based on task requirements, equipment characteristics, and operating conditions.

\subsection{Industrial Agent System Layer}
The Industrial Agent System Layer comprises six modules: memory, knowledge and mechanisms, tools, interaction, orchestration, and verification. These systems equip the industrial agent with task context, domain knowledge, external capabilities, and execution constraints. They form the agent's core composition and are used throughout the operational process described in the runtime layer.

\subsubsection{Memory System}
The memory system stores the current task context and information retained from previous tasks. Short-term memory records task objectives, recent observations, intermediate results, and tool outputs. Long-term memory preserves equipment histories, fault cases, and operational experience. Episodic memory records specific events and task trajectories, while semantic memory stores facts and concepts learned from these experiences. Memory entries are associated with equipment identifiers, timestamps, and operating conditions. These associations help the agent retrieve experience relevant to the current task and distinguish records from different equipment states or production periods. For example, a previous fault case is useful only when its equipment configuration and operating conditions are considered.

\subsubsection{Knowledge and Mechanism System}
The knowledge and mechanism system contains the domain knowledge and physical principles used by the agent. Knowledge graphs represent industrial entities and their relationships, including equipment topology, process dependencies, and fault associations. Technical documents, operating procedures, and engineering rules provide further information about equipment capabilities and process requirements. Physical mechanisms describe relationships among process variables and specify operating limits. The system makes these relationships available as equations, rules, or references to mechanism-based models in the foundation layer. The agent can use them to interpret observations, examine possible causes, and assess proposed actions. Knowledge retrieval can be restricted to the equipment, process, and operating conditions involved in the current task.

\subsubsection{Tool System}
The tool system provides interfaces to external software, models, data services, and industrial equipment. APIs and Model Context Protocol (MCP) services expose capabilities such as database queries, model inference, simulation, and access to engineering or manufacturing software. Each tool has a defined function, input schema, output format, and access permissions. Industrial device interfaces connect the agent to equipment through existing control or service systems. These interfaces specify the available operations and their execution conditions. Commands affecting physical processes remain subject to the equipment's control logic and safety interlocks. The tool system therefore defines what the agent can access and how requests must be expressed.

\subsubsection{Interaction System}
The interaction system provides natural language and multimodal interfaces between the agent and its users. Users can describe tasks, supply engineering documents or images, and ask questions about equipment or production processes. The system associates these inputs with the relevant task and industrial objects. The same interfaces present results, supporting evidence, execution status, and requests for clarification or authorization. Engineers can correct task assumptions, revise constraints, or approve proposed operations. User feedback is recorded with the task context so that subsequent actions reflect the revised requirements.

\subsubsection{Orchestration System}
The orchestration system organizes skills, tools, and agents for task execution. A skill can combine models, knowledge, and tool calls into a reusable procedure for a task such as fault diagnosis or production scheduling. The system records the required inputs, dependencies, and outputs of these procedures. Workflows, task graphs, and state machines specify how the available capabilities are connected. They define execution order, conditional branches, resource assignments, and handoffs between agents. Established industrial procedures can be encoded as fixed steps, with designated points for agent analysis or human review. The runtime layer uses these structures to carry out plans and coordinate participating agents.

\subsubsection{Verification System}
The verification system checks proposed outputs and actions against physical mechanisms, operational constraints, and safety and compliance requirements. It uses the knowledge and mechanism system as a source of equations, rules, and limits. Checks may cover output structure, parameter ranges, equipment states, process dependencies, and consistency with physical relationships. Simulation or specialized models can be used when an action's consequences require further assessment. Verification results determine whether a proposal meets the specified execution conditions or requires revision or human review. Actions requiring human authorization are held until approval is recorded. The system also retains the checked proposal, supporting evidence, and verification result for later inspection.

\subsection{Industrial Agent Runtime Layer}
The Industrial Agent Runtime Layer describes the agent's operation during a task. The process follows a loop of perception and understanding, planning, reasoning and decision-making, action, and feedback learning. Multi-agent collaboration supports tasks that require several agents. At each stage, the agent uses the models in the foundation layer and the systems in the system layer. Observations and execution results update the task context and may prompt another cycle or a revision of the current plan.

\subsubsection{Perception and Understanding}
Perception and understanding establish the current state of the industrial environment and the meaning of the task. Inputs include sensor measurements, equipment alarms, production logs, inspection images, engineering drawings, and user instructions. The agent associates these inputs with the relevant equipment, processes, and time periods, accounting for differences in sampling rates and data availability. It then interprets the observations using task context and domain knowledge. This stage identifies operating conditions, available resources, and events that may affect the task. Missing or conflicting observations can prompt additional data queries or requests for clarification before planning proceeds.

\subsubsection{Planning}
Planning translates the task objective into a sequence of activities. The agent decomposes the objective into subtasks and identifies their dependencies, required resources, and completion conditions. For industrial tasks, the sequence must respect process order, equipment availability, and production schedules. The agent can use established procedures from the orchestration system or construct a task sequence from available skills and tools. The resulting plan specifies what needs to be done and in what order. Execution feedback may require changes to individual steps or a revision of the plan when equipment states, resource availability, or task requirements change.

\subsubsection{Reasoning and Decision-Making}
Reasoning and decision-making determine how the planned activities should be carried out under current conditions. The agent examines candidate actions using observations, retrieved experience, industrial knowledge, and physical mechanisms. Causal, temporal, and topological relationships help explain equipment behavior and assess dependencies between processes. Specialized models can provide numerical predictions, while industrial world models can estimate the consequences of candidate actions. The agent compares the alternatives against objectives such as quality, efficiency, cost, energy consumption, and equipment health. It selects an action within the applicable constraints or revises the plan if no feasible option is found.

\subsubsection{Action}
Action carries out the selected operation through the tool system. The agent prepares the tool inputs, submits the request for verification, and obtains human authorization where required. Operations may include querying a database, running a simulation, updating a production schedule, or submitting an equipment operation request. Actions affecting physical processes pass through existing control or service interfaces and remain subject to equipment interlocks and control limits. The agent records the returned results and execution status. A failed request or an unexpected equipment response can interrupt the sequence and trigger further observation or replanning.

\subsubsection{Feedback Learning}
Feedback learning evaluates task outcomes and uses the results to improve subsequent behavior. The agent compares the observed outcome with the task objective and, where available, the predicted effect of its actions. Evaluation considers task completion and relevant industrial indicators, such as product quality, production efficiency, energy consumption, and equipment condition. Task trajectories, successful procedures, and failure cases are retained in memory for later use. Repeated discrepancies can indicate that a strategy, model, or operating assumption needs revision. The accumulated experience informs later planning and supports the refinement of operating conditions within verified limits. Updated observations then enter the next cycle of perception and understanding.

\subsubsection{Multi-Agent Collaboration}
Multi-agent collaboration coordinates tasks that span equipment, processes, or engineering disciplines. Agents divide the work according to their roles and capabilities, share relevant observations and intermediate results, and coordinate execution. Task allocation accounts for dependencies between subtasks and competition for shared resources. During execution, agents exchange progress updates and report changes that affect other participants. Conflicting proposals must be reconciled against the shared objective and operational constraints before dependent actions proceed. For example, a maintenance agent may report that a machine is unavailable, prompting a manufacturing agent to revise its schedule. Collaboration therefore operates across the runtime loop as task conditions evolve.

\subsection{Industrial Agent Application Layer}
The Industrial Agent Application Layer covers five groups of agents: R\&D, manufacturing, testing, management, and maintenance. Each group uses the underlying models, systems, and runtime processes for tasks in its domain. Agents access the engineering software, business systems, and equipment services required by these tasks. Activities that cross domain boundaries may involve several agents, such as revising a production schedule after an equipment fault or checking manufacturing constraints during product design.

\subsubsection{R\&D Agents}
R\&D agents assist engineers throughout requirements analysis, engineering design, simulation, and design optimization. They interact with CAD, CAE, EDA, PLM, and other engineering systems to analyze requirements, generate design alternatives, optimize parameters, review drawings, and conduct simulation studies. For complex products involving multiple engineering disciplines, design decisions need to account for both engineering principles and downstream manufacturing constraints. Different agents can therefore collaborate on related design activities and refine solutions as the design evolves.

\subsubsection{Manufacturing Agents}
Manufacturing agents support scheduling, process execution, resource coordination, and quality and logistics management. They use information about orders, process routes, equipment availability, materials, and work in progress to prepare or revise production plans. Equipment failures, material shortages, and quality deviations may require changes to task sequences or resource assignments. Agents can coordinate these changes through MES, MOM, and related production systems. During process execution, they track progress, identify deviations from the plan, and request corrective actions. Equipment operations are carried out through authorized interfaces under the existing control logic.

\subsubsection{Testing Agents}
Testing agents support test planning, test execution, and result validation. They translate test objectives, product specifications, and applicable standards into procedures, parameter settings, and acceptance criteria. During execution, they interact with simulation platforms or test equipment to collect measurements and monitor test conditions. Unexpected observations may prompt additional measurements or a review of the test setup. The agents compare results with acceptance criteria and use physical models or historical data to investigate discrepancies. Test records retain the configuration, measurements, and analysis needed to review the conclusions.

\subsubsection{Management Agents}
Management agents support procurement, supply chain coordination, inventory management, and finance. They analyze orders, stock levels, supplier information, production requirements, and cost records through enterprise systems such as ERP and WMS. Their tasks include identifying material shortages, comparing procurement options, checking delivery commitments, and assessing budget deviations. Decisions that affect several business functions require coordination with the relevant agents and personnel. For example, a proposed procurement change may need to be checked against production schedules, inventory limits, and financial approval rules.

\subsubsection{Maintenance Agents}
Maintenance agents support condition monitoring, anomaly detection, fault diagnosis, and remaining useful life prediction. They examine sensor measurements, alarm logs, inspection results, and maintenance records to assess equipment health. Diagnostic models and knowledge about equipment topology and fault propagation help identify possible causes of abnormal behavior. Prognostic models estimate degradation trends and remaining useful life under specified operating assumptions. The agents use these findings to recommend inspections or maintenance activities, considering equipment criticality and production requirements. After maintenance, they compare equipment behavior with its previous condition and record the outcome for future diagnosis and planning.

\section{Industrial Applications}

\newcommand{\todo}[1]{\textcolor{red}{#1}}
This section focuses on the full lifecycle of industrial products and discusses typical application scenarios and potential implementation methods for five categories of agents: R\&D and design, production and manufacturing, testing and evaluation, business management, and operations and maintenance services.

\subsection{R\&D Agents}

R\&D and design encompass phases such as requirements analysis, conceptual design, detailed design, and simulation verification. R\&D and design-oriented intelligent agents are software-based and deeply integrated with toolchains such as CAD, CAE, and EDA, supporting multiple rounds of design iterations from requirements to final solutions.

\subsubsection{Requirements Analysis and Proposal Evaluation.}
The agent analyzes lengthy design briefs, breaks down abstract requirements into a system of quantifiable metrics, retrieves historical proposals and design specifications to generate multiple candidate conceptual proposals, and outputs a trade-off analysis. This can be implemented using retrieval-enhanced generation and a multi-agent review mechanism.

\subsubsection{Generative Design and CAD Modeling.}
Using natural language descriptions or sketches as input, this process models parts by generating parametric modeling scripts or directly creating 3D models~\cite{khan2024text2cad}, and iterates in conjunction with specialized algorithms such as topological optimization and strength verification to achieve a rapid closed-loop cycle from requirements to geometry to performance.

\subsubsection{Automated CAE Simulation.}
Agents drive simulation software to complete the entire workflow, including geometry cleanup, meshing, boundary condition setup, solution monitoring, and result interpretation. For example, upon receiving a task to improve the cooling efficiency of turbine blades, the agent autonomously modifies the parameters of the cooling hole layout and drives the CFD solver to perform multiple rounds of simulation-based optimization.

\subsubsection{Chip and Electronic Design.}
Agents generate EDA toolchain scripts and autonomously orchestrate design workflows, such as ChatEDA~\cite{wu2024chateda}, which enables end-to-end EDA automation, and ChipNeMo~\cite{liu2023chipnemo}, which is based on domain-adapted large language models.

\subsubsection{Materials and Formulation R\&D.}
By combining literature-based knowledge mining, experimental plan formulation, and automated experimental platforms, these agents enable closed-loop exploration of material compositions and formulations. In the field of chemistry, Coscientist~\cite{boiko2023coscientist} and ChemCrow~\cite{bran2024chemcrow} have demonstrated the feasibility of such agents autonomously designing and executing experiments.

\subsection{Manufacturing Agents}

Manufacturing agents take the form of a combination of software and hardware, deeply integrated with MES, APS, SCADA, and production line equipment to address demand fluctuations and on-site disruptions in high-variety, low-volume production.

\subsubsection{Dynamic Scheduling and Dispatching}

The agent monitors the real-time status of orders, equipment, and work-in-progress. In the event of rush orders or equipment failures, it generates rescheduling plans based on process routes and delivery deadline constraints and dispatches them to the MES for execution. This can be achieved through a multi-agent coordination mechanism involving orders, equipment, and materials~\cite{zhou2021marlscheduling}.

\subsubsection{Process Parameter Optimization}

By fusing process mechanism models with historical production data, the system provides online recommendations and performs closed-loop adjustments to parameters such as welding current, injection molding temperature, and cutting parameters, balancing efficiency and energy consumption while meeting quality constraints.

\subsubsection{Quality Inspection and Root Cause Analysis}

After a vision model detects a defect, the agent automatically correlates process parameters, equipment status, and batch information to perform root cause analysis, generating analysis reports and remediation recommendations to achieve a closed-loop process from detection to attribution and remediation.

\subsubsection{Embodied Flexible Operations}

By combining the task planning capabilities of large language models with robotic operational skills, natural language commands drive robotic arms and AGVs to perform flexible operations such as assembly and material handling. This approach can draw on embodied intelligence technology frameworks such as SayCan~\cite{ichter2023saycan}, PaLM-E~\cite{driess2023palme}, and RT-2~\cite{zitkovich2023rt2}.

\subsection{Test Agents}

Test and evaluation agents combine the functions of software and intelligent test equipment, covering the entire process from test plan design to execution control and data analysis.

\subsubsection{Test Plan Generation}

Automatically generates test outlines based on evaluation metrics and product characteristics; uses Design of Experiments (DOE) methods to optimize test point layout; and reduces the number of tests and the test cycle while ensuring evaluation coverage.

\subsubsection{Test Execution and Safety Control}

These agents drive test benches to automatically execute procedures according to the outline while monitoring multiple sensor parameters in real time. They trigger safety actions, such as shutdown protection, in response to abnormal conditions like overheating or excessive rotational speed, and dynamically adjust subsequent test points based on interim results.

\subsubsection{Test Case Generation}

For industrial software and embedded systems, test cases are automatically generated from requirements documents and source code, and regression testing is executed on Software-in-the-Loop (SiL) and Hardware-in-the-Loop (HiL) platforms~\cite{xue2024llm4fin,dakhel2024mutap,shin2013hil}.

\subsubsection{Data Analysis and Report Generation}

The system automatically performs test data preprocessing, feature extraction, and compliance assessment of performance metrics; generates structured test reports; and identifies potential anomalies and performance degradation trends by comparing data with historical test records.

\subsection{Management Agents}

Business management agents are purely software-based and integrate with enterprise systems such as ERP, SCM, and CRM through a natural language interface (NL2SQL) to support cross-domain business decision-making.

\subsubsection{Supply Chain Collaboration and Simulation}

The agents continuously monitor supply chain risk events, conduct what-if simulations, and generate contingency plans. OptiGuide~\cite{li2023optiguide} demonstrates an approach that combines large language models with optimization solvers to address supply chain what-if analysis questions; multi-agent systems can further simulate negotiation games among suppliers, manufacturers, and logistics providers.

\subsubsection{Demand Forecasting and Planning}

By integrating historical orders, market signals, and macroeconomic data, the system generates demand forecasts and coordinates rolling adjustments to the master production schedule and material requirements planning.

\subsubsection{Business Analysis}

It performs metric queries, attribution analysis, and report generation through conversational interactions, automatically transforming open-ended questions about issues such as a decline in gross profit margin in a certain region into multi-step data queries and attribution reasoning chains, as demonstrated by ChatBI~\cite{lian2024chatbi}.

\subsubsection{Contract and Compliance Review}

It identifies risks in contract terms, assists in the preparation of bidding and tender documents, and conducts compliance checks, while consolidating knowledge of corporate policies and processes to build organizational memory.

\subsection{Maintenance Agents}

Operations and maintenance service agents exist in both software and hardware forms. They are designed for equipment health management and after-sales service, integrating multimodal state sensing with knowledge of failure mechanisms.

\subsubsection{Fault Diagnosis and Root Cause Analysis}

By integrating multimodal signals such as vibration, acoustics, infrared, and log data, the agent autonomously invokes signal analysis and diagnostic tools to pinpoint the root cause of failures and generate remediation recommendations. It can draw on the tool invocation paradigms of RCAgent~\cite{wang2024rcagent} and D-Bot~\cite{zhou2024dbot} in the AIOps field.

\subsubsection{Predictive Maintenance}

Based on equipment health assessments and remaining-life predictions, it automatically generates maintenance schedules and coordinates spare parts inventory and maintenance resource scheduling. For example, it can continuously track the performance degradation trajectory of individual aircraft engines to dynamically optimize decisions regarding engine replacement and repair.

\subsubsection{Embodied Inspection}

Inspection robots or drones equipped with vision-language models autonomously plan inspection routes in scenarios such as substations and chemical plants, identify meter readings, leaks, and abnormal noises, and automatically generate inspection reports and alert work orders.

\subsubsection{Maintenance Assistance and Intelligent Customer Service}

The agent builds a knowledge-based Q\&A service based on maintenance manuals and historical work orders, combines this with augmented reality (AR) to provide remote maintenance guidance, and enables the automatic routing and intelligent dispatch of after-sales service work orders.

\section{Future Prospects}
\subsection{Industrial Domain Knowledge-Driven Multimodal Foundation Models}

Industrial intelligent agents are expected to evolve from general-purpose foundation models toward multimodal foundation models driven by industrial domain knowledge. These models will integrate heterogeneous industrial data, including technical documents, engineering designs, sensor time series, inspection images, simulation data, control programs, and operational records, to establish unified representations of equipment, processes, and production tasks. General-purpose foundation models, industrial multimodal models, time-series foundation models, and specialized task models are expected to evolve collaboratively. Industrial mechanisms, process rules, and control logic will also be increasingly incorporated into model training and inference, enabling a transition from correlation-based learning toward reasoning and decision-making constrained by domain knowledge and physical mechanisms.

\subsection{Industrial World Model-Driven Agents}

Industrial world models will provide an important foundation for prediction, planning, and policy validation in industrial intelligent agents. Future industrial world models are expected to evolve from models of individual devices and local processes toward dynamic representations covering personnel, equipment, materials, processes, and production environments. By integrating digital twins, industrial mechanism models, generative simulation, and data-driven modeling, these world models will enable agents to predict system evolution and evaluate the potential consequences of different actions. They will also provide virtual environments for agent training, policy validation, and continuous optimization.

\subsection{Software and Hardware Collaborative Industrial Intelligent Agents}

Industrial intelligent agents will increasingly integrate industrial software, edge computing platforms, and physical equipment to establish closed-loop interaction among perception, decision-making, execution, and feedback. Future system architectures are expected to combine intelligent decision-making with deterministic industrial control in a layered manner, improving task understanding and dynamic planning while preserving real-time performance, operational constraints, and safety guarantees. As intelligent capabilities move closer to production environments, cloud-edge-device collaboration, model compression, low-latency inference, and specialized computing hardware will become increasingly important. These developments will facilitate the deeper integration of intelligent agents into robots, industrial equipment, production lines, and other physical systems.

\subsection{Trustworthy Industrial Intelligent Agents with Long-Term Memory}

Industrial intelligent agents will gradually develop long-term memory mechanisms organized around industrial entities such as equipment, products, materials, processes, and tasks. Such mechanisms will enable the continuous accumulation of equipment histories, abnormal events, process changes, operational experience, and execution trajectories, thereby supporting long-horizon tasks and knowledge reuse. Long-term memory will also be integrated with domain knowledge, real-time data, and tool-generated results. Through mechanisms such as neuro-symbolic reasoning, evidence association, and execution trace recording, industrial intelligent agents will achieve improved interpretability, verifiability, and traceability of critical decisions.

\subsection{Industrial Intelligent Agents with Endogenous Security}

As industrial intelligent agents acquire stronger capabilities for autonomous decision-making, tool use, and physical execution, their security risks will extend beyond conventional data and network security to include model hallucination, knowledge contamination, unauthorized access, tool misuse, cascading errors, and unsafe physical operations. Future security mechanisms are expected to be embedded throughout the entire agent lifecycle, covering data, models, knowledge, memory, decision-making, and execution. Robustness enhancement, zero-trust architectures, and trusted computing technologies will jointly support proactive risk identification, dynamic access control, critical action verification, and abnormal recovery. Such mechanisms will enable industrial intelligent agents to achieve higher levels of autonomy while maintaining safety, controllability, and traceability.

\subsection{Evaluation of Industrial Intelligent Agents}

The evaluation of industrial intelligent agents is expected to evolve from isolated model-level testing toward system-level evaluation covering complete task processes and real-world operational outcomes. Future evaluation frameworks should comprehensively assess perception and understanding, task planning, tool use, execution performance, collaboration, reliability, safety, and computational efficiency. Particular attention should be given to task success rates, operational stability, and exception-handling capabilities in long-horizon and multi-step tasks. Digital twins, virtual factories, hardware-in-the-loop environments, and real production systems can jointly provide hierarchical testing environments. Combined with repeatable and comparable benchmark tasks and evaluation metrics, these environments will support capability assessment, technological iteration, and the large-scale deployment of industrial intelligent agents.

\section{Conclusion}
{
This article systematically reviewed LLM-driven industrial agents from the perspectives of key challenges, conceptual evolution, taxonomy, system architecture, industrial applications, and future directions. A four-layer architecture was presented to organize foundation models, agent systems, runtime mechanisms, and application services for closed-loop industrial task execution. Representative applications across R\&D, manufacturing, testing, management, and maintenance were further discussed. Overall, industrial agents are expected to drive the transition from isolated industrial intelligence toward collaborative, adaptive, and reliable industrial autonomy.
}

\bibliographystyle{ieeetr}
\bibliography{reference}

\end{document}